\documentclass[sigconf]{acmart}
\renewcommand\footnotetextcopyrightpermission[1]{} 
\usepackage{booktabs} 
\usepackage{placeins}

\acmDOI{10.1145/nnnnnnn.nnnnnnn}

\acmISBN{978-x-xxxx-xxxx-x/YY/MM}

\acmConference[GECCO '18]{the Genetic and Evolutionary Computation
Conference 2018}{July 15--19, 2018}{Kyoto, Japan}
\acmYear{2018}
\copyrightyear{2018}

\acmArticle{4}
\acmPrice{15.00}

\begin{document}

\title{Functional Generative Design: \\An Evolutionary Approach to 3D-Printing}

\copyrightyear{2018} 
\acmYear{2018} 
\setcopyright{acmcopyright}
\acmConference[GECCO '18]{Genetic and Evolutionary Computation Conference}{July 15--19, 2018}{Kyoto, Japan}
\acmBooktitle{GECCO '18: Genetic and Evolutionary Computation Conference, July 15--19, 2018, Kyoto, Japan}
\acmPrice{15.00}
\acmDOI{10.1145/3205455.3205635}
\acmISBN{978-1-4503-5618-3/18/07}

\author{Cem C. Tutum}
\affiliation{
	\institution{The University of Texas at Austin}
}
\email{tutum@cs.utexas.edu}

\author{Supawit Chockchowwat}
\affiliation{
	\institution{The University of Texas at Austin}
}
\email{chockchowwatsc@utexas.edu}

\author{Etienne Vouga}
\affiliation{
	\institution{The University of Texas at Austin}
}
\email{evouga@cs.utexas.edu}

\author{Risto Miikkulainen}
\affiliation{
	\institution{The University of Texas at Austin,\\and Sentient Technologies, Inc.}
}
\email{risto@cs.utexas.edu}

\renewcommand{\shortauthors}{C. C. Tutum et. al.}

\begin{abstract}

Consumer-grade printers are widely available, but their ability to print complex objects is limited. Therefore, new designs need to be discovered that serve the same function, but are printable. A representative such problem is to produce a working, reliable mechanical spring. The proposed methodology for discovering solutions to this problem consists of three components: First, an effective search space is learned through a variational autoencoder (VAE); second, a surrogate model for functional designs is built; and third, a genetic algorithm is used to simultaneously update the hyperparameters of the surrogate and to optimize the designs using the updated surrogate. Using a car-launcher mechanism as a test domain, spring designs were 3D-printed and evaluated to update the surrogate model. Two experiments were then performed: First, the initial set of designs for the surrogate-based optimizer was selected randomly from the training set that was used for training the VAE model, which resulted in an exploitative search behavior. On the other hand, in the second experiment, the initial set was composed of more uniformly selected designs from the same training set and a more explorative search behavior was observed. Both of the experiments showed that the methodology generates interesting, successful, and reliable spring geometries robust to the noise inherent in the 3D printing process. The methodology can be generalized to other functional design problems, thus making consumer-grade 3D printing more versatile.



\end{abstract}



\maketitle



\section{Introduction}

\begin{figure}
\includegraphics[width=3.0in]{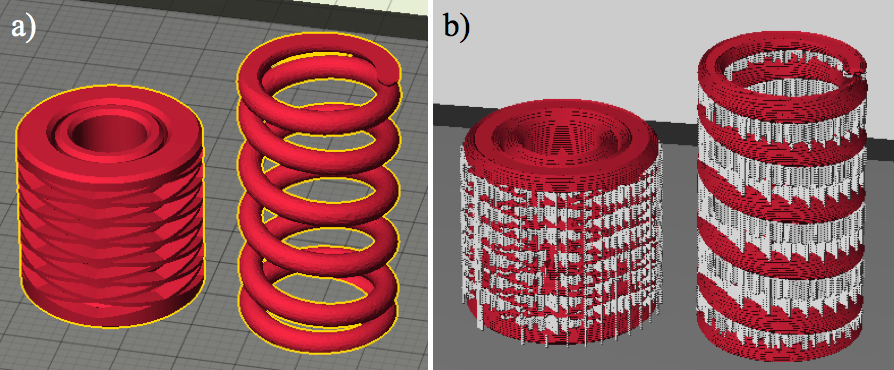}
\caption{a) CAD models of two different types of springs are shown, i.e. a coil spring designed in one piece (on the left), and a standard helical spring (on the right). b) The 3D printing models are prepared (material indicated with red color) and support structures (extra material depicted with grey color) are added in the spring models . This extra material needs to be removed.}
\label{fig_helical_springs}
\end{figure}

Fused Deposition Modeling (FDM), commonly known as 3D Printing, is a practical prototyping technique in which a 3D digital model of a design is sliced via dedicated software into thin layers that are fused on top of each other to form the final product. FDM is widely adopted for consumer-gade printers because it is simple and inexpensive. It is an extrusion-based 3D printing process where the heated and liquefied thermoplastic material has to be laid down in layers to be supported by the layers beneath them. Thus, the maximum slope of overhanging geometry becomes a significant limitation in the case of printing intricate objects \cite{Yu2017}. Most common way to solve this problem is the use of extra support structure composed of the same printing material, which is removed after the printing process. Besides the waste of material, the post-processing is time-consuming and frustrating, especially when the support structures are printed in difficult-to-access regions or extra surface treatment operations (such as sanding or acetone vapor smoothing) are needed. Most importantly, there is a risk of breaking delicate parts of the printed object during the removal process. This would be an important limitation, for instance in case of printing a helical spring Fig.\ref{fig_helical_springs}, which is known to perform well in designs where propelling or suspension capabilities are needed. Thus, new functional products need to be designed taking the 3D printing-related constraints into account.

Topology optimization is one way of designing stuructures subject to manufacturability constraints \cite{Zegard16_topopt}. Due to the complexity of the resulting organic-looking shapes, it mostly attracted academic researchers rather than practitioners or manufacturers. The gap between the topology optimization and its real world applications has recently been closing with the advances in 3D printing. However, designing structures which have to withstand a variety of loading conditions, and to perform certain functional purposes, as well as to comply with manufacturing constraints (tolerances, etc.) is still a challenging problem due to computational requirements in topology optimization. The model resolution should be high and gray-scale regions in the resulting design should be minimized to get it manufactured without the loss of information.
One potential solution to handle computationally intensive or time consuming optimization problems is to use surrogate or (low-fidelity) meta-models \cite{Forrester09}. These approximation functions learn the mathematical mapping from the variable space to the function space using a relatively small sample size. Most known surrogates in the literature vary from simple polynomial regression models and moving least squares to neural networks, radial basis functions, Kriging, and support vector regression. Despite the variety in their mathematical construction, they all work based on the same consecutive principles: \textit{training} (learning) and \textit{testing} (prediction or generalization). The trained model allows the user to predict any response at an unknown design set at a negligible cost.
In recent years, unsupervised learning, in particular generative design algorithms such as Variational Autoencoders (VAE) and Generative Adversarial Networks (GAN) have become more popular in computer graphics and 3D modeling. In \cite{Umetani2017}, Umetani proposes a transformation method of 3D meshes into parametric models, which enables user to manually control the variables learned by an autoencoder, for exploring other possible designs. With an expanded version of a standard autoencoder, authors \cite{Nash2017} show the capability of VAE in fitting the distribution of complex 3D model space. In addition to presenting the flexibility of VAE, the authors further investigate the blurriness of surface reconstruction, suggesting VAE's capability to encode 3D global structure. On the other hand, \cite{Li2017} develops a recursive neural network, automatically assembled with an autoencoder components and trained with a GAN. Towards a goal to map the generative structure manifold, the designed neural architecture is able to encode both local geometry and hierarchical structure of 3D models. All these works aim at both reconstructions of latent variables to 3D models as well as exploring authentic designs from limited examples, however the functional aspects of the design are overlooked. This paper is presented as an attempt to reduce this gap. As a proof of concept, a car-launcher mechanism is used as a test domain, however the methodology can be generalized to other functional design problems as well.

\begin{figure}
\includegraphics[width=2.5in]{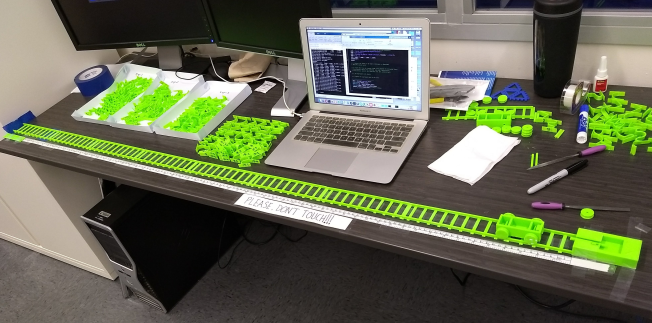}
\caption{The green parts located along the edge of the table constitute the 3D printed car-launcher mechanism. The track has a length of 157cm and a ruler is atached next to the track for the measurements. A link to the video of the experimental setup can be found in \cite{CarLauncherVideo}.}
\label{fig_exp_setup1}
\end{figure}
\begin{figure}
\includegraphics[width=2.5in]{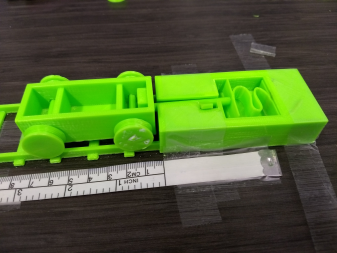}
\caption{Closer view of the car-launcher mechanism is shown above. A test spring is loaded and ready to propel the car.}
\label{fig_exp_setup2}
\end{figure}

The organization of the paper is as follows: First, the subcomponents of the methodology, i.e. the VAE model, fitness evaluation, evolutionary surrogate and Efficient Global Optimization algorithms, are introduced. Next, experimental setup as shown in Fig.\ref{fig_exp_setup1} and \ref{fig_exp_setup2}, is described in detail, and then, quantitative results are accompanied by the 3D prints of the evolved spring designs. Finally, outcomes of the methodology is briefly discussed and some ideas for the future work are elaborated.

\section{Methodology}

\begin{figure}
\includegraphics[width=3.25in]{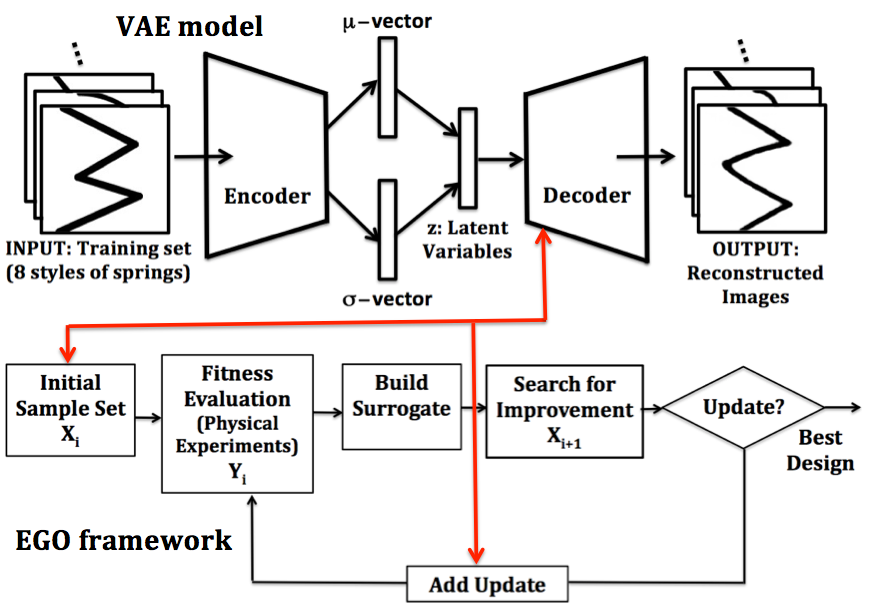}
\caption{Overall framework is mainly composed of the VAE and EGO algorithms. The decoder of the trained VAE model is used for generation of spring models for the optimum candidates (arrays of LVs) during EGO iterations.}
\label{fig_framework}
\end{figure}

Overall methodology (see Fig.\ref{fig_framework}) briefly involves a Variational Autoencoder (VAE), a 3D printer to produce the springs and an experimental setup for evaluating spring desings and Efficient Global Optimization (EGO) algorithm. The main driver of the pipeline is the EGO algorithm (see Section \ref{sec_EGO}), which uses a surrogate (i.e. function approximation) tuned to have a correlation between design parameters (i.e., encodings in this case) and fitness values. VAE is a generative model, simply meaning that it can generate new instances looking like those in the training set (see Section \ref{sec_VAE}). Initially, a training set of images containing spring-like objects, as combination of straight lines, arcs and Bezier-curves, have been prepared to train the VAE. Next, the initial sample set for the EGO, to build the first surrogate model, is prepared. A standard Genetic Algorithm (GA) is used to tune the hyperparameters of the surrogate. Next, the VAE decoder is applied to convert the list of encoded spring desings into 3D models. All 3D spring models in the initial set are 3D printed and tested in the car-launcher mechanism to get their fitness scores assigned. Since the physical experiments are noisy and in some cases result in missing values (see Section \ref{sec_fitness}), i.e. a regression type of Kriging model (as opposed to the general use of interpolating Kriging) is preferred as a surrogate function to drive the EGO iterations. Same GA is also employed here to search for the maximizer of the Expected Improvement (EI) function, where the improvement is defined as finding a function value better than the current best one. After an infill (candidate) design vector is found at the end of an iteration, it is converted into a 3D model and printed for the fitness evaluation. This procedure is repeated for a certain number of iterations. As a result, a list of interesting, reliable and functional spring designs are evolved.

\FloatBarrier

\subsection{Training data for the VAE model} \label{sec_VAEtraining_set}

\begin{figure}
\includegraphics[width=3.25in]{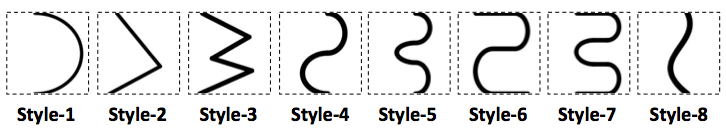}
\caption{Training set contains 8 spring styles (10,000 samples of each).}
\label{fig_training_set}
\end{figure}

In order to avoid the need for the support structure, \textbf{1)} Two dimensional (2D) spring-like geometries were randomly created as black-and-white images (see Fig.\ref{fig_training_set}) with the resolution of 150x150, \textbf{2)} Black regions are smoothened (i.e., by tracing the bitmap to a vector), \textbf{3)} Finally, these smoothened curves are extruded in the third dimension (depth) for creating the 3D model (i.e., STL file) followed by the use of the slicing software where the printing instructions are produced (model is discretized). Printed models are then tested in the experimental setup and their fitness values are assigned to each individual. 

Some of the training samples are shown in Fig. \ref{fig_training_set}. Eight intuitive designs, here they are referred to "styles", are determined: \textbf{Style-1}: Single-Arc, \textbf{Style-2}: Single-ZigZag, \textbf{Style-3}: Double-ZigZag, \textbf{Style-4}: Double-Arc, \textbf{Style-5}: Tripple-Arc, \textbf{Style-6}: Double-Arc-with-Lines, \textbf{Style-7}: Tripple-Arc-with-Lines, and \textbf{Style-8}: Bezier curve. Each style had 10,000 randomly created samples, where the main shape of the corresponding style is preserved while thickness, relative position of different segments are varied. 

\subsection{VAE model} \label{sec_VAE}

VAE is used to learn proper representation space of the spring designs as well as reduce dimensionality and generate new likely designs \cite{Kingma2014}. The goal of learning is to approximate posterior probability $p_{\theta}(\mathbf{z} | \mathbf{x})$ with recognition model $q_{\phi}(\mathbf{z} | \mathbf{x})$. In this work, VAE is consists of 2 convolutional layers followed by a fully connected layer (to mean $\mu$ and $\log{\sigma^2}$, which is used to sample $z$ from $\mathcal{N}(\mu, \sigma^2)$) and a fully connected layer (from latent $\mathbf{z}$) followed by 2 transposed convolutional layers in the encoder and decoder respectively. The cost function at training is defined as the sum of a Kullback-Leibler divergance term $p_{\theta}(\mathbf{z})$ to $q_{\phi}(\mathbf{z} | \mathbf{x})$ and a reconstruction term.

\begin{eqnarray}
\left.\begin{aligned}
& \mathcal{L}(\phi, \theta; \mathbf{x}) = \mathcal{D}_{KL}(q_{\phi}(\mathbf{z} | \mathbf{x}) || p_{\theta}(\mathbf{z})) + \mathbb{E}_{\mathbf{z} \sim q_{\phi}(\mathbf{z} | \mathbf{x})} \left[ \log{p_{\theta}(\mathbf{x} | \mathbf{z})} \right] \\
& \mathcal{D}_{KL}(q_{\phi}(\mathbf{z} | \mathbf{x}) || p_{\theta}(\mathbf{z})) = \frac{1}{2} \sum_{j} \left( 1 - \mu_j^2 + \log{\sigma_j^2} - \sigma_j^2 \right)
\end{aligned}\right.
\end{eqnarray}

Since the spring design bitmap $\mathbf{x}$ arguably represents the occupancy of material at each spatial coordinate, the output is assumed to be Bernoulli and leads to cross entropy reconstruction error

\begin{equation}
\mathbb{E}_{\mathbf{z} \sim q_{\phi}(\mathbf{z} | \mathbf{x})} \left[ \log{p_{\theta}(\mathbf{x} | \mathbf{z})} \right] = \sum_{i} \left( x_i \log{y_i} + (1 - x_i) \log{1-y_i} \right)
\end{equation}



\begin{figure}
\includegraphics[height=3in, width=3in]{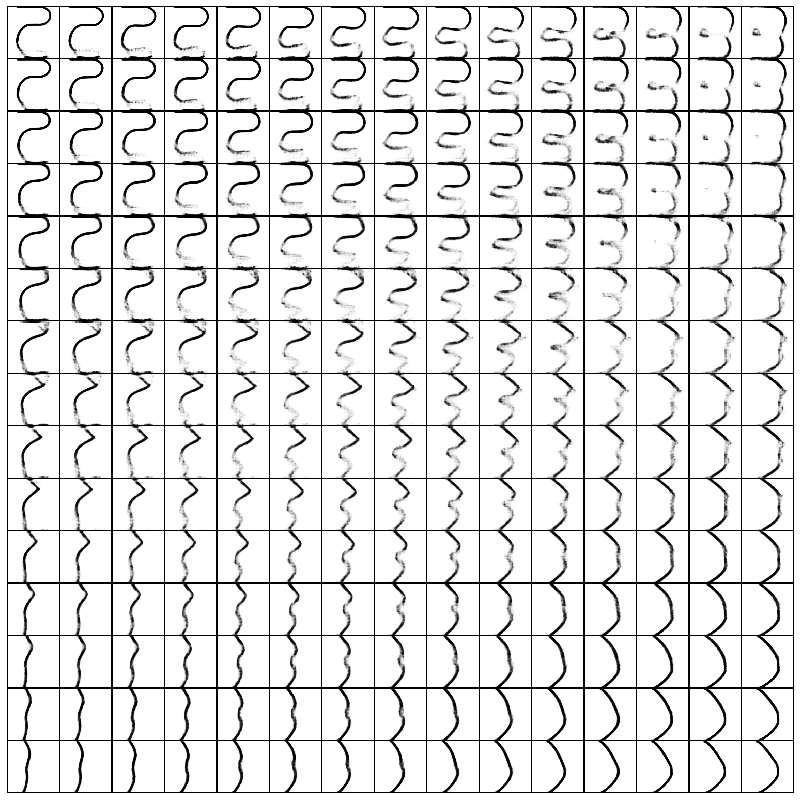}
\caption{Resulting spring designs obtained via interpolation in the 12D-latent variable space.}
\end{figure}

\begin{figure}
\includegraphics[height=2.4in, width=3.5in]{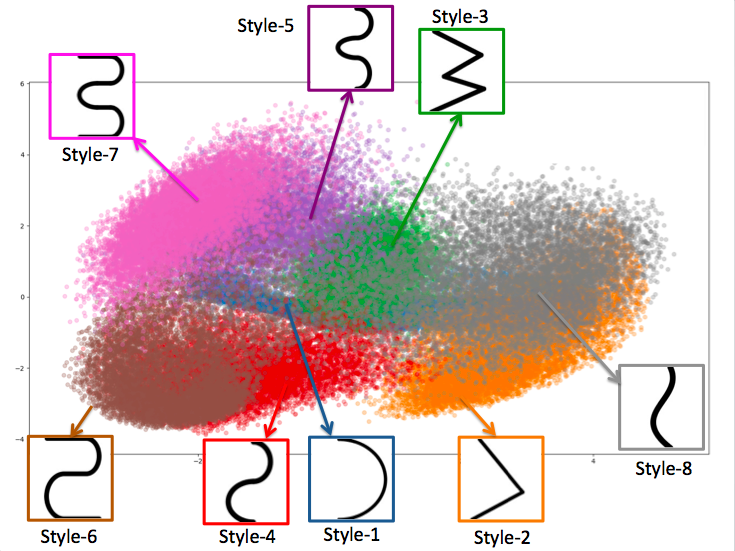}
\caption{The distribution of spring designs in the latient space (i.e., LV-3 versus LV-1) colored according to their spring styles.}
\end{figure}

Given that the VAE gives a real-numbers decoded bitmap and the products possibly create disconnections which are wasteful to be printed, since they do not attach to the two bases of a spring. A simple bitmap traversal filtering is employed to discretize pixels and remove unwanted part by tracking those pixels with values greater than a fixed threshold and collectively form a connected component with both bases along neighbors (up, down, left, and right pixels). Note that this also results in a blank design if the decoded does not contain such connected component. Although some designs visually suggest a curve but barely miss a gap in between, the filtering will disregard and eliminate the whole designs. More work can be done to recover such case for a denser design space, but it might as well complicate patterns for an optimizer.

\subsection{Fitness evaluation} \label{sec_fitness}
\begin{figure}
\includegraphics[width=3.25in]{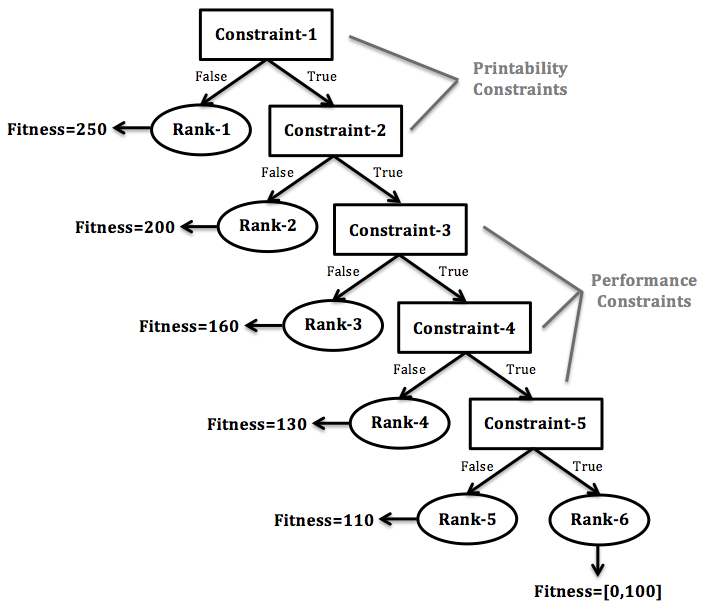}
\caption{A tree structure is defined for ranking the performance of individuals with respect to several criteria and assigning their fitness values. Only the Rank-6 solutions are feasible and all other solutions ranked from 5 to 1, categorized as infeasible, are assigned gadually increasing fitness values to guide the search in the infeasible search region.}
\label{fig_ranking_tree}
\end{figure}

The car-launcher mechanism is used as a test bed for evaluating the performance of the springs. Springs are 3D printed with two additional flat surfaces at both ends for a stable loading step. A cylindrical bar is pushed against the top flat surface of the spring along the cylindrical groove of the launcher. With the release of the spring, the car is pushed along the rails. A video of the experimental setup is shown at the provided link \cite{CarLauncherVideo}. The goal is to design a spring to propel the car to a distance of 75 cm in a consistent and reliable way. This is a more challenging and less intuitive task than maximizing the travelling distance of the car, as it requires identifying the stiffest possible spring design which remains soft enough to be loaded into the launcher. Moreover, the elastic behavior of the spring plays a crucial role in the current selected problem, because relatively stiff springs tend to have worse fatigue performance, and on the other hand, the soft springs cannot store enough energy to push the car to the desired distance. The overall success of a spring design is formulated in the following equation,

\begin{equation}
f(x) = \sum_{i=1}^{n_{exp}} MSE_i =  \frac{1}{10} \sum_{i=1}^{10} |d_i-75|^2,
\label{eqn_fitness}
\end{equation} 
where Mean Square Error (MSE) is minimized using the 12D-vector of latent variables as the design parameters. Each spring is tested 10 times. The designs are also subject to some constraints which can be categorized into two types (see Fig.\ref{fig_ranking_tree}): \textbf{1)} Printability constraints, \textbf{2)} Performance constraints. Printability constraints have two sub-categories: \textbf{C1)} Is the design (i.e., its image form) blurry, or in other words, could VAE decoder produce a visible pattern? \textbf{C2)} If there is any visible pattern, is it connected all the way from the top to the bottom surface? Performance constraints, which satisfy the printability constraints, are further divided into three sub-categories: \textbf{C3)} Is the spring loadable (i.e., not too stiff)? \textbf{C4)} Did the spring work through all 10 experimental repeatitions without breaking? \textbf{C5)} Did the car stay on track in at least 5 out of 10 experiments? It should be noted here that if the car reaches to the end of the rail system or if it flips over due to excessive unstability, it is recorded as two penalties (i.e. $NaN$ value is assigned) and the spring design is accepted as infeasible in case of having more than 5 $NaN$ values.

As aforementioned, each of the 3D printed springs were tested 10 times in the car-launcher mechanism, distance measurements for the fitness assignment and constraint violations (i.e. ranking) were recorded. Feasible (Rank-6) solutions were scaled into [0,100]-scale, where 0 represents the best fitness value (i.e., which indicates that the car frequently stops in the near vicinity of 75 cm) and 100 represents the worst feasible value. Following the fitness assignment of feasible solutions, non-printable (blurry or disconnected) designs are assigned Rank-1 (fitness=250) and Rank-2 (fitness=200), whereas printable, but poor performing designs (very stiff, fragile or non-stable) were assigned Rank-3 (fitness=160), Rank-4 (fitness=130) and Rank-5 (fitness=110). Via ranking system, missing values, which indicate regions of the search space to be avoided, are easily identified. Moreover, the constrained problem is converted into an unconstrained problem where only one surrogate model for the fitness function needed to be built.


\subsection{Surrogate (Regressing Kriging)}  \label{sec_Kriging}
Interpolating Kriging is a well-known surrogate technique that is frequently used to approximate computationally expensive functions in the course of optimization \cite{Krig51, Sacks89}. The procedure starts with obtaining a sample data (i.e., \textit{n}-design sets each having \textit{d}-variables), $\mathbf{X_{n \times d}}= [\mathbf{x}^{(1)}, \mathbf{x}^{(2)},..., \mathbf{x}^{(n)}]^T$, and a corresponding vector of scalar responses $\mathbf{y_{n \times 1}}= [y^{(1)}, y^{(2)},..., y^{(n)}]^T$. In this work, n=48, and d=12 as the size of the output (i.e. LV space) of the VAE encoder. Any two response values in the sample set are correlated with the Gaussian basis function, as a function of the absolute distance between the sample points,

\begin{equation}
cor\left[ y(\mathbf{x}^{(i)}), y(\mathbf{x}^{(j)}) \right]= \prod_{k=1}^{d} exp \left( -\theta_k  \left|\mathbf{x}_k^{(i)}-\mathbf{x}_k^{(j)} \right|^2  \right),
\label{eqn_correlation}
\end{equation}

where $\theta_k$ is a correlation parameter or hyperparameter (i.e., $\theta_k = \theta_1, \theta_2, ..., \theta_d$) which controls how fast the correlation changes from one point to the other one along each dimension. Eq. \ref{eqn_correlation} is used to build the symmetric correlation matrix (\textbf{R}) of all \textit{n}-points in \textbf{X}, which will be used in the process of tuning the unknown hyperparameter $\theta_k$ using maximum likelihood estimation (MLE). However, since the response values (i.e., fitness values assigned after the spring tests) are noisy, a regression model is preferred. Therefore, to filter noise, a regression constant $\lambda$ needs to be tuned as an extra hyperparameter together with $\theta_k$'s.

It is well-known that the hyperparameter space for the Kriging model is non-convex, therefore use of a global search algorithm is advised in general \cite{Toal2008}. Thus, a real parameter (floating number) Genetic Algorithm (rGA) is used for searching the Regressing Kriging hyperparameters (total of 13 parameters, i.e. $\theta_{1\rightarrow12}\:and\:\lambda$). The rGA has a tournament selection with a tournament size of 3. It employs a blend crossover with a crossover probability of $p_{cx}=90\%$ and  $\alpha=0.9$, where $\alpha$ controls the extent of the interval in which the new values can be drawn for each attribute on both side of the parents' attributes. The rGA also uses a Gaussian mutation of mean $\mu=0.0$ and standard deviation $\sigma=0.01 \: |UB_i-LB_i|$ on the input individual (i.e., $1\%$ of each variable bounds).

After tuning the Kriging model parameters, the next step is to predict a new response value, i.e., a fitness function value, at an unobserved design vector, \textbf{x*}, using the sample data that are used to train the Kriging model. Regressing Kriging predictor ($\hat{y}$) has such a form,

\begin{equation}
\hat{y}(\mathbf{x}^*) = \hat{\mu} + \mathbf{r}^T (\mathbf{R} + \lambda \mathbf{I})^{-1} \left( \mathbf{y}-\mathbf{1}\hat{\mu} \right),
\label{eqn_predictor}
\end{equation}
where $\mathbf{r}$ is the linear vector of correlations between the unknown point to be predicted ($\mathbf{x}^*$) and the known sample points ($\mathbf{x}$), $\hat{\mu}$ is the estimated mean, and $\mathbf{1}$ is an unit vector of size n x 1. Kriging is in general known for its good performance in fitting complex functional behavior; however, what makes Kriging a very popular surrogate method is in essence its ability to estimate the \textit{mean squared error} (MSE) at the unknown point,

\begin{equation}
\hat{s}^2(\mathbf{x}^*) = \hat{\sigma}^2 \left[ 1+\lambda-\mathbf{r}^T (\mathbf{R}+\lambda \mathbf{I})^{-1} \mathbf{r}+\frac{1-\mathbf{1}^T (\mathbf{R}+\lambda \mathbf{I})^{-1}\mathbf{r}}{\mathbf{1}^T (\mathbf{R}+\lambda \mathbf{I})^{-1}\mathbf{1}} \right],
\label{eqn_MSE}
\end{equation}
where $\hat{s}^2$ represents the MSE estimate. However, since $\hat{s}$ is not zero at sample points due to noise as is not the case in interpolating Kriging, there is a risk of resampling during the EGO iterations (see Section \ref{sec_EGO}). To avoid this, $\hat{s_{ri}}$ (re-interpolation error) is used instead,

\begin{equation}
\hat{s_{ri}}^2(\mathbf{x}^*) = \hat{\sigma_{ri}}^2 \left[ 1-\mathbf{r}^T \mathbf{R}^{-1} \mathbf{r}+\frac{1-\mathbf{1}^T \mathbf{R}^{-1}\mathbf{r}}{\mathbf{1}^T \mathbf{R}^{-1}\mathbf{1}} \right],
\label{eqn_MSEri}
\end{equation}
where $\hat{\sigma_{ri}}^2$ is the modified variance,

\begin{equation}
\hat{\sigma_{ri}}^2 = \frac{(\mathbf{y}-\mathbf{1}\hat{\mu})^T(\mathbf{R}+\lambda\mathbf{I})^{-1}\mathbf{R}(\mathbf{R}+\lambda\mathbf{I})^{-1}(\mathbf{y}-\mathbf{1}\hat{\mu})}{n}.
\label{eqn_variance_ri}
\end{equation}

\subsection{Efficient Global Optimization} \label{sec_EGO}
Knowing the fact that the Kriging model just constructed on the limited number of sample points (\textit{initial sample set}) is only an approximation for the underlying black-box function; thus, new sample points (\textit{infill points}) should iteratively be sought to update, or in other words, to improve the accuracy of the surrogate. This update procedure (infill criterion) can balance both exploration and exploitation purposes, i.e., simultaneously utilizing the information of the predictor $\hat{y}$(x) calculated by Eq. \ref{eqn_predictor} and the estimation of the variance $\hat{s_{ri}}^2$(x) calculated by Eq. \ref{eqn_MSEri}. Jones et al. \cite{Jones98} suggested an algorithm called \textit{Efficient Global Optimization} (EGO), which relies on building iteratively a probabilistic model (i.e., Kriging, section \ref{sec_Kriging}) of the objective function and a criterion based on improving upon the best sample found so far, $y_{best}$, by searching this probabilistic model. Recall that the Kriging predictor is the realization of a Gaussian process Y(\textbf{x}) with the mean $\hat{y}$ and the variance $\hat{s}^2$(\textbf{x}); therefore, due to the uncertainty in the predictor, an improvement at a point \textbf{x} can be defined as,

\begin{equation}
I(\mathbf{x}) = max\left( y_{best}-Y(\mathbf{x}) \right) ,
\label{eqn_improvement}
\end{equation}
which can be used to maximize the expectation of it (\textit{expected improvement}) as the infill criterion (\cite{Sasena02,Viana09}),

\begin{equation}
\begin{split}
E[I(\mathbf{x}]) = & (y_{best}-\hat{y}(\mathbf{x})) \Phi\left( \frac{y_{best}-\hat{y}(\mathbf{x})}{\hat{s}(\mathbf{x})} \right) + \\
 & \hat{s}(\mathbf{x}) \phi\left( \frac{y_{best}-\hat{y}(\mathbf{x})}{\hat{s}(\mathbf{x})}\right),
\label{eqn_expected_improvement}
\end{split}
\end{equation}
where $\Phi$(.) and $\phi$(.) are the \textit{cumulative distribution function} and the \textit{probability density function} of a normal distribution, respectively. Readers are referred to \cite{Viana11} for the derivation of Eq. \ref{eqn_expected_improvement}. The search space for the expected improvement is highly multimodal as in the case of hyperparameter tuning of the Kriging model, therefore the same rGA algorithm mentioned in the previous section is applied here to find the global optimizer of the EI, Eq.\ref{eqn_expected_improvement}. This candidate  solution is then fed back to the surrogate again to simultaneously improve the accuracy of the approximation as well as to get closer to the global optimum. The general framework of the EGO is shown in Fig.\ref{fig_framework} (bottom). EGO iterates until a user-defined stopping criterion is met, e.g., total number of infill points, change in the objective function, tolerance on MSE, etc.


\section{Results}
In this section, results of the proposed functional generative design methodology are presented. Two different experiments are employed and their performances are tested on the car launcher mechanism. Both of the initial sets of the EGO are composed of 36 solutions of which 16 of them are selected from the training set for the VAE model and 20 of them are selected with respect to the Gaussian distribution since the VAE model also fits the efficient representation in the same probability distribution. In Experiment-1, those 16 solutions from the training set are selected in random, keeping in mind that some of the spring styles might have not been represented in the initial set at all while some are sampled more than once. The initial set in Experiment-2 differs in a way that two samples from each spring style are chosen to increase the diversity in the initial sample set. In addition to the difference in selection of the first 16 solutions (i.e., random vs. 2 samples per each style), in Experiment-1, only the EGO-candidate solution (i.e., maximizer of Eq.\ref{eqn_expected_improvement}) is printed and tested for its fitness value. However, in Experiment-2, in addition to the EGO-candidate solution, 3 more solution vectors are numerically perturbed around it with $sigma=0.05$ and a total of 4 update solutions are printed and evaluated every infill iteration. The following two sections show detailed results about the two experiments.

\subsection{Experiment-1} \label{sec_Exp1}

\begin{figure}
\includegraphics[width=3.25in]{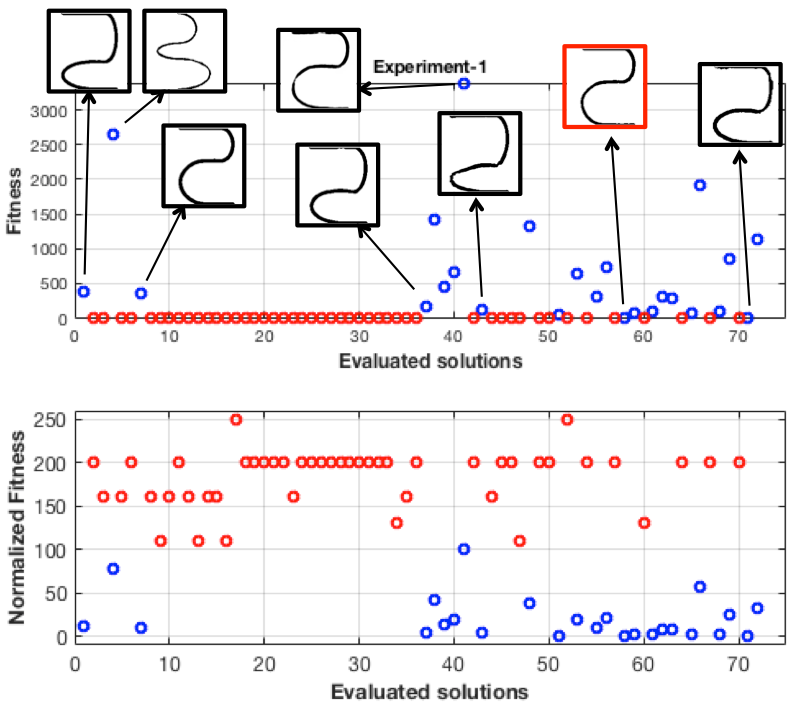}
\caption{Fitness and Normalized Fitness values of the designs evaluated in Experiment-1 are indicated with blue (feasible) and red (infeasible) markers. The optimum spring design, Design-58 (Style-6), is indicated with the red box.}
\label{fig_exp1_runs}
\end{figure}

Fig.\ref{fig_exp1_runs} (top) shows the fitness values (Eq.\ref{eqn_fitness}) in the vertical axis and the sequencially evaluated solution indices in the x-axis (i.e., total of 72 design evaluations). The blue marker indicates the feasible solutions and the red marker depicts the infeasible solutions where either of Rank-1 through 5 are violated. In the top graph, since the fitness of the infeasible solutions cannot be computed, zero value is assigned to their fitnesses. However, in order to differentiate them, their normalized fitness values, see Fig.\ref{fig_exp1_runs} (bottom), are assigned according to the ranking system introduced in Section-\ref{sec_fitness}, so non-printable designs have either the normalized fitness of 250 (blurry designs) or 200 (disconnected designs), whereas those violating performance constraints take normalized fitness values of 160 (very stiff, not even loadable), 130 (fragile, broken during tests) or 110 (not stable, out of track more than 5 times). All other solutions marked with blue color are assigned normalized fitness values between 0 and 100, where 0 and 100 indicate the best and worst feasible solutions, respectively.

According to the normalized fitness values, the best two designs (Design-58 and 71, respectively, modified Style-6) is found on the second and the last infill iterations of the EGO. In the initial set, there were one Styles-1-3 and 4, three Style-5, four Style-6, two Style-7 and four Style-8 spring designs were chosen randomly and only three of them were feasible. Thus Style-2 was not present in the initial set and Styles-1-3 and 4 were each represented by only one sample. Out of those three feasible designs, Style-6 had the best fitness value in the initial set. In the next infill iterations of EGO, an exploitative search behavior can be observed from the small boxes of design images added in Fig.\ref{fig_exp1_runs} (top). The best design had propelled the car almost constantly to the target distance of 75cm. Designing such a spring, which performs almost the same ten times in a row and achieves a precise goal, is not an easy task. The big difference in performance with respect to almost invisible changes in the design space (i.e., latent variable space) and printing instructions shown in Fig.\ref{fig_exp1_runs} proves the efficiency of the proposed method. 

\begin{figure}
\includegraphics[width=3.25in]{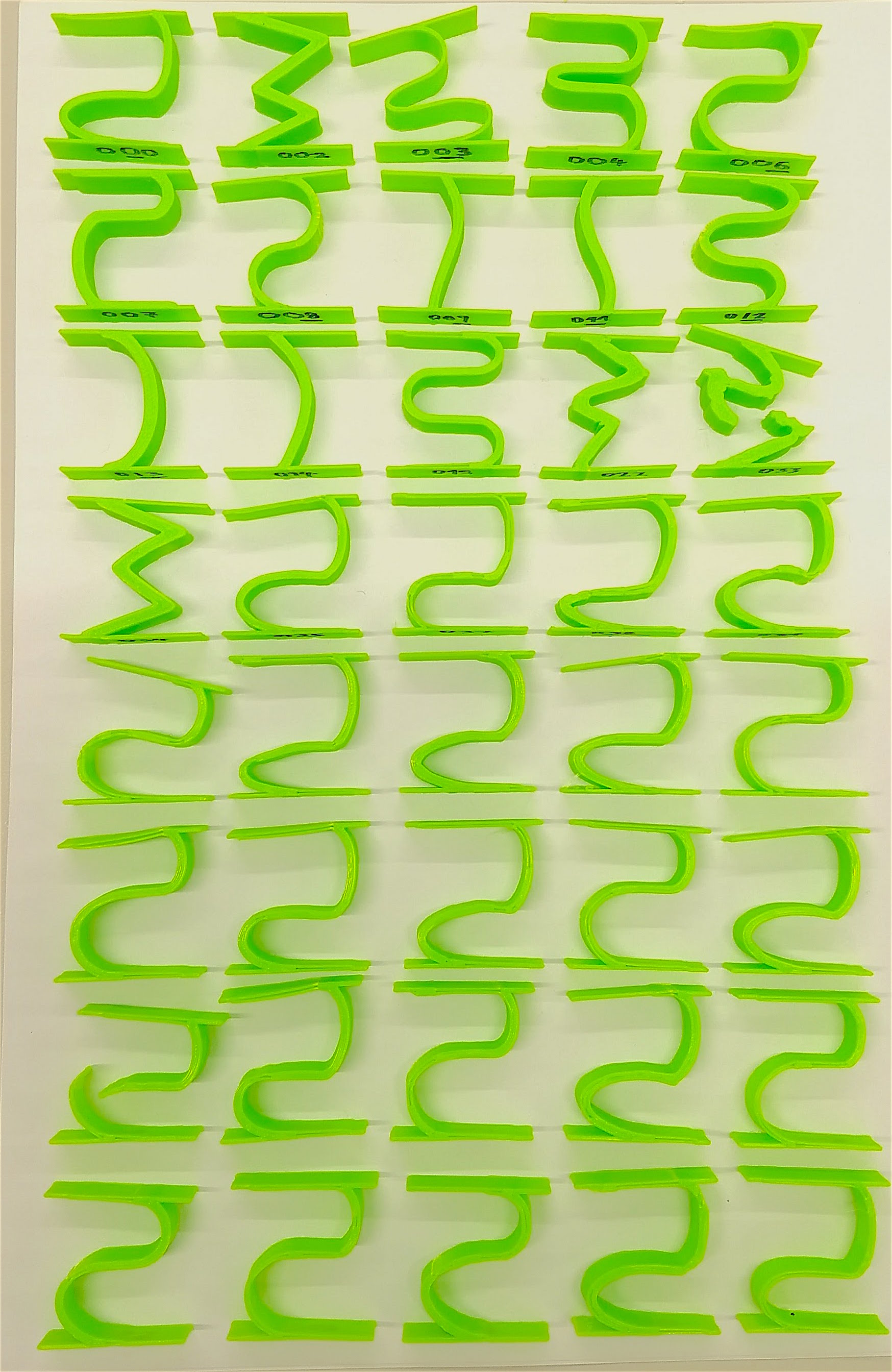}
\caption{All of the printable spring designs tested in Experiment-1 are shown above. There are two broken springs which are ranked as 4, also visible in Fig.\ref{fig_exp1_runs} (bottom).}
\label{fig_Exp1_allprints}
\end{figure}

Fig.\ref{fig_Exp1_allprints} shows all the printed and tested designs (Rank-3 through 6 solutions) during Experiment-1. Optimum design is the first one in the sixth row.

\subsection{Experiment-2}  \label{sec_Exp2}

\begin{figure}
\includegraphics[width=3.25in]{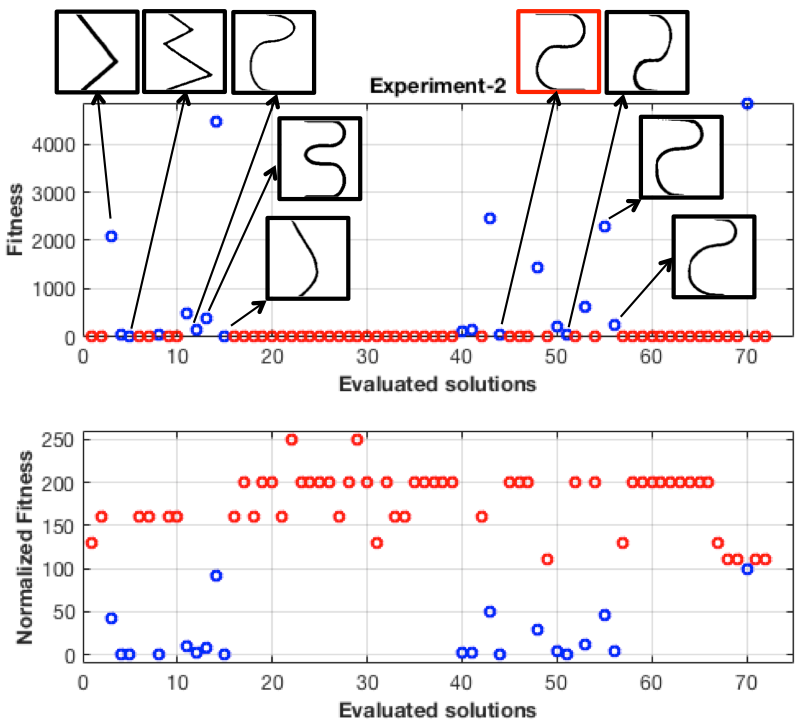}
\caption{Fitness and Normalized Fitness values of the designs evaluated in Experiment-2 are indicated with blue (feasible) and red (infeasible) markers. The optimum spring design, Design-43 (Style-6), is indicated with the red box. However, Design-15 (Style-8), the fifth small image on top figure, also has a similar fitness value.}
\label{fig_exp2_runs}
\end{figure}

\begin{figure}
\includegraphics[width=3.25in]{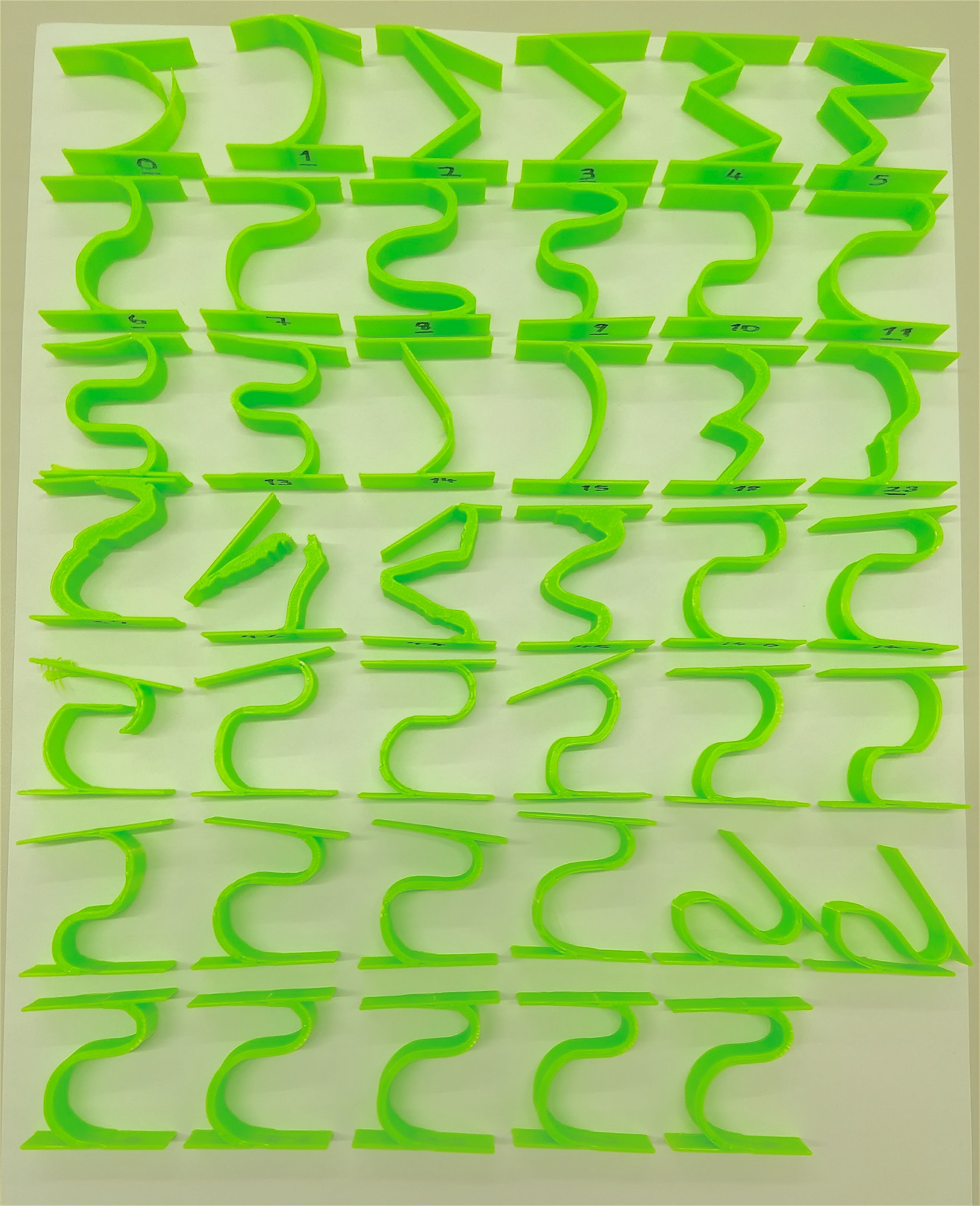}
\caption{All of the printable spring designs tested in Experiment-2 are shown above. There are four broken springs which are ranked as 4, also visible in Fig.\ref{fig_exp2_runs} (bottom)}
\label{fig_Exp2_allprints}
\end{figure}

Fig.\ref{fig_exp2_runs} (top and bottom) show the fitness and normalized fitness values as similar to Fig.\ref{fig_exp1_runs}. According to the normalized fitness values, the best two designs (Design-43 and 15, respectively, Style-6 and 8) is found on the second infill iteration and the initial set of the EGO. The initial set was composed of more uniformly selected (two samples from each spring style) designs from the same set of samples that was used for training the VAE model. In the next infill iterations of EGO, more explorative search behavior is observed as referred to the printed spring designs in Fig.\ref{fig_Exp2_allprints}. For instance, there are designs which look like the mixtures of Style-3 (Double-ZigZag) and Style-5 (Tripple-Arc) as well as Style-8 (Bezier) at the last two springs of the third row (see Fig.\ref{fig_Exp2_allprints}). Such explorative pattern is repeated  in the following 2 or 3 EGO iterations, then Style-6 spring started to emerge as the optimum solution as also seen in Experiment-1. The performance of the best design in Experiment-2 is almost as good as the one in Experiment-1, however the first one had more time to evolve due to exploitative search observed in Experiment-1. 

\section{Discussion and Future Work}

The results show that a surrogate-based global search algorithm, coupled with a real parameter GA (rGA) and a problem-specific constraint handling methodology, can be very effective at generating complex, reliable and functional designs despite a very noisy nature of the fitness landscape, having various sources of uncertainty, and a budget of limited number of function evaluations. Two experiments performed in this study clearly indicate how succesfully the method can produce new spring designs by controlling a fewer number of design features learned by a VAE model. The car-launcher mechanism is used as proof of concept here, which can be extended into various other more complex design problems such as in compliant mechanisms, aero-dynamic surfaces or structural components. The proposed methodology improves the reliability of a design by integrating manufacturing (e.g. 3D printing) instructions indirectly in the discretized digital model.\\
Despite the success of the methodology, there are a couple of promising improvements to consider. First of all, the performance of the VAE model can be improved by a more dedicated hyperparameter optimization of both the network topology and training parameters. A lower dimensional latent variable space would be easier to approximate with a surrogate as well as to search with an optimization algorithm. Second most important improvement could be achieved with a finer level of evaluations of Rank-1 (blurriness) and especially Rank-2 (disconnected). Currently all Rank-1 and 2 solutions assigned 250 and 200, respectively, which does not help differentiating within each correnponding style-cluster. Variation within a cluster would help directing the search from infeasible region to boundaries of the feasible designs. This would also reveal even more potentially interesting designs in the course of optimization, because new style infill designs being strictly evaluated as infeasible (not the level of infeasibility) don't have time to evolve further since the search is almost immediately directed away from the potential region. \\
While "finer grading" could be a solution to direct the search away from the disconnected design regions, an alternative approach would be to repair such designs (e.g., to naively connect the ends with straight lines), however it should be noted that such a quick fix would deviate from the original idea of learning efficient representations of feasible designs. A better repair method could be, inspired from the K-Means clustering algorithm, to find the closest spring pattern in that particular latent variable space to fill out the gaps.\\
Fully automating the generative procedure with a simulator would enable more detailed sampling and infill evaluations. The increased search capacity would also make it possible to find multiple promising designs and time to fine tune their parameters. However development of a physical model for such a nonlinear problem is not an easy task and could also be prohibiting even for a surrogate-based method due to high computational demands. A Co-Kriging model \cite{Toal2011}, which combines a set of physical experiments (i.e., a high-fidelity model) and a relatively simple FE model (i.e., a low-fidelity model) or even a rigid-body beam model, could be iteratively evolved within the EGO framework to reduce the need for a high-fidelity model.

\section{Conclusions}
This paper proposes a methodology for designing functional 3D-printed springs, using a car-launcher mechanism as proof of concept. The methodology consists of three components: First, a low-dimensional search space is learned through a variational autoencoder; second, a surrogate model is built to correlate the learned representations with the physical designs; and third, a genetic algorithm is used to simultaneously update the hyperparameters of the surrogate and to optimize an infill criterion to evolve functional and reliable desings. Two experiments were then performed: First, the initial set of desings were selected randomly, thus the non-uniform representation of each different spring design evolved more exploitative search characteristics. On the other hand, the latter approach had more explorative characteristics due to a denser and more uniform representation of the feasible designs as a head start. Both of the experiments showed that the methodology generates interesting, successful, and reliable spring geometries despite the noise inherent in the 3D printing process.

\section{Acknowledgments}

The authors wish to acknowledge the funding and support provided
by the Freshman Research Initiative Program at the College of
Natural Sciences in The University of Texas at Austin.


\bibliographystyle{ACM-Reference-Format}
\bibliography{sample-bibliography} 

\end{document}